\documentclass[sigconf,nonacm]{acmart}
\usepackage[ruled,vlined,linesnumbered]{algorithm2e}

\usepackage{amsmath}
\usepackage{graphicx}
\usepackage{subcaption}
\newcommand{\algname}[0]{\texttt{Astraea}\xspace}

\usepackage{caption}
\usepackage{hyperref}
\begin{document}

\title{Astraea: A State-Aware Scheduling Engine for LLM-Powered Agents}

\author{Hongqiu Ni}
\affiliation{
  \institution{University of Science and Technology of China}
  \country{China}
}
\author{Jiabao Zhang}
\affiliation{
  \institution{University of Science and Technology of China}
  \country{China}
}
\author{Guopeng Li}
\affiliation{
  \institution{University of Science and Technology of China}
  \country{China}
}
\author{Zilong Wang}
\affiliation{
  \institution{University of Science and Technology of China}
  \country{China}
}
\author{Ruiqi Wu}
\affiliation{
  \institution{University of Science and Technology of China}
  \country{China}
}

\author{Chi Zhang}
\authornote{Corresponding author.}
\affiliation{%
  \institution{Hefei University of Technology}
  \country{China}
}

\author{Haisheng Tan}
\authornote{Corresponding author.}
\affiliation{%
  \institution{University of Science and Technology of China}
  \country{China}
}

\begin{abstract}

Large Language Models (LLMs) are increasingly being deployed as intelligent agents. Their multi-stage workflows, which alternate between local computation and calls to external network services like Web APIs, introduce a mismatch in their execution pattern and the scheduling granularity of existing inference systems such as vLLM. Existing systems 
typically focus on per-segment optimization which prevents them from minimizing the end-to-end latency of the complete agentic workflow, i.e., the global Job Completion Time (JCT) over the entire request lifecycle. 
To address this limitation, we propose \algname, a service engine designed to shift the optimization from local segments to the global request lifecycle. \algname employs a state-aware, hierarchical scheduling algorithm that integrates a request's historical state with future predictions. It dynamically classifies requests by their I/O and compute-intensive nature and uses an enhanced HRRN policy to balance efficiency and fairness. \algname also implements an adaptive KV cache manager that intelligently handles the agent state during I/O waits based on the system memory pressure. Extensive experiments show that \algname reduces average JCT by up to 25.5\% compared to baseline methods. Moreover, our approach demonstrates strong robustness and stability under high load across various model scales.
\end{abstract}

\keywords{LLM Serving Systems, LLM Inference Engine, Agentic workflows, Latency Optimization}

\maketitle
\section{Introduction}
\label{sec:intro}

The widespread deployment of Large Language Models (LLMs) as cloud services, accessed via Web APIs, has catalyzed a new generation of powerful web applications. Meanwhile, LLMs are gradually evolving from language processors into intelligent agents (LLM Agents) capable of autonomous planning and task execution.
These agents leverage external tools~\cite{ref1, ref2, ref3} by programmatically issuing a series of \textbf{API calls} to web services and other resources, enabling them to accomplish more complex tasks, such as autonomously browsing the web~\cite{ref4, ref5, ref6}, solving issues on GitHub~\cite{ref7, ref8, ref9}, and proving challenging mathematical problems~\cite{ref10, ref11}. 

\begin{figure}[htbp]
    \centering
        \includegraphics[width=0.82\columnwidth]{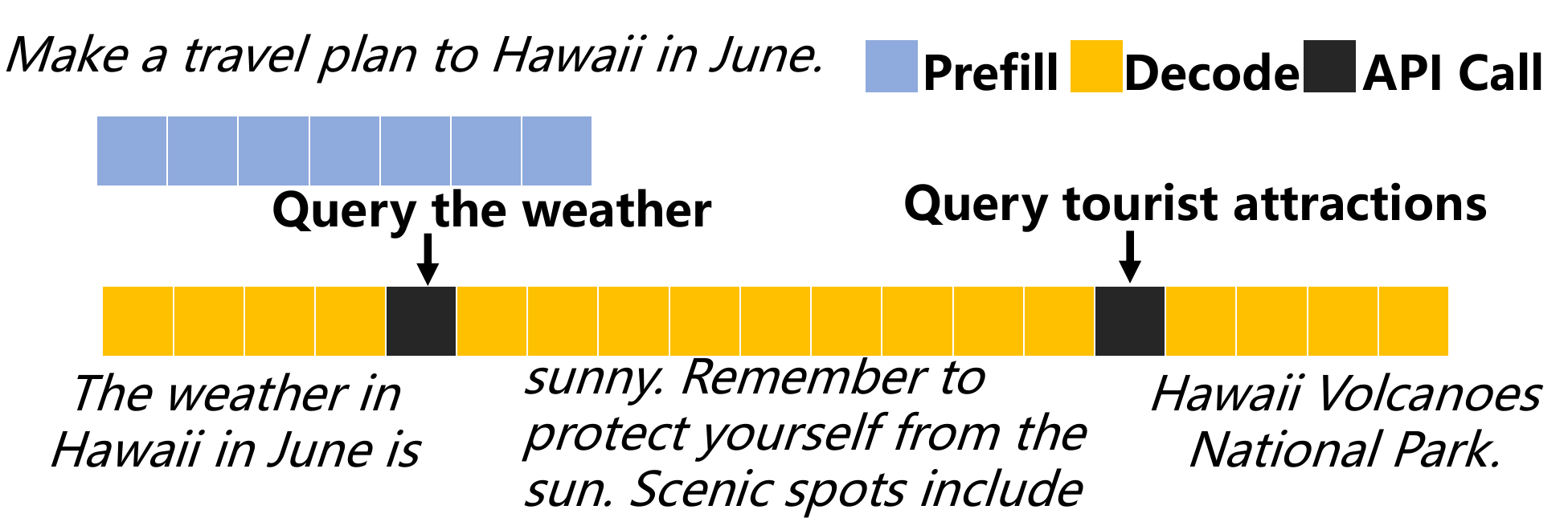} 
        \caption{An example of an agentic inference workflow.}
        \label{fig:augLLM}
        \Description{This is an example image showing ...}
\end{figure}

The execution of these agentic tasks introduces a multi-stage inference workflow, which is structurally different from the process of traditional LLMs.
A standard inference process involves two distinct stages: a compute-intensive \textbf{Prefill} stage that processes the input context in parallel, followed by a memory-bound \textbf{Decode} stage that generates tokens autoregressively. Throughout this process, the \textbf{key-value (KV) cache} is employed to store intermediate states, thereby accelerating the generation of subsequent tokens.
However, an agent executes a more complex \textbf{agentic workflow} composed of multiple such inference processes, as illustrated in the example in Figure~\ref{fig:augLLM}. When a user issues a request (e.g., ``Make a travel plan to Hawaii in June.''), the system executes the first inference process, which continues until the agent generates a special token to trigger an API call to query the weather. At this point, the LLM computation pauses, and the system enters an I/O wait state. Once the API returns a result (e.g.,``(The weather is) sunny.''), this new information is appended to the context, and the process resumes with another Prefill and Decode cycle. This evolution from a single computational task into a dynamic, multi-stage workflow of alternating computation and external API calls introduces unprecedented challenges for the underlying service infrastructure, demanding a rethinking of how to schedule requests and manage resources for these advanced systems.
\addvspace{-2mm}

This new agentic workflow exposes two flaws in existing LLM serving systems. First, current schedulers~\cite{ref13,ref14} are largely unaware to the duration of API calls, leading to severe Head-of-Line (HoL) blocking. These systems only focus on modeling the computation stages, for instance, $S^3$~\cite{ref15} predicts decode length to improve GPU utilization and accelerate inference. However, in LLM Agent scenarios, decode time is often a negligible fraction of the request's total lifecycle. For example, an API call to SerpAPI for web search can last several seconds, while the decode phase is mere milliseconds. A scheduler which ignores dominant API duration cannot make effective resource allocation or scheduling decisions.
Second, current systems like vLLM treat each segment of agentic workflow as an independent task. This approach optimizes for a local metric—the response time of the current segment—but fails to optimize the end-to-end Job Completion Time (JCT) for the entire multi-stage request. These dual shortcomings reveal a critical research gap and we need overcome two challenges to design a new paradigm filling this gap:

\noindent\textbf{Challenge 1: The High-Stakes Trade-off in State Management.} Agentic workflows introduce long I/O wait periods where the large KV cache remains idle on the GPU. This presents a critical resource management dilemma: preserving the cache in GPU memory reduces system throughput by occupying scarce resources, while discarding it incurs significant recomputation latency upon resumption. This creates an inherent 
trade-off between the single-request latency and system-wide throughput. 

\noindent\textbf{Challenge 2: Lifecycle-Aware Scheduling for Global Optimization.} The local properties of a segment are severely decoupled from its global impact on the end-to-end JCT, as the latter is determined by future I/O phases. This disconnection introduces the risk of HoL blocking, where a decision to prioritize a seemingly optimal segment can starve other requests during a long API call duration, leading to disastrous global consequences. 


To address the above challenges, we propose a novel scheduling paradigm, to unify a request's historical state with the prediction of its future behavior, and present \algname, a lifecycle-centric inference service engine for LLM agentic workflows, which is to optimize the global Job Completion Time (JCT) rather than myopic per-segment metrics. Our main contributions are summarized as:

\begin{itemize}
\item We formalize the scheduling problem for LLM agent workflows and prove it to be NP-hard, providing a theoretical foundation for heuristic algorithm design (Sec.~\ref{sec:problem_formalization}).  
\item We design a novel state-aware hierarchical preemptive scheduling mechanism, named \algname, which unifies historical workflow behavior and future predictions. At the macro level, requests are dynamically classified according to their computing and I/O characteristics; at the micro level, scheduling order is determined by a combination of the waiting time and predicted service duration, balancing efficiency and fairness  (Sec.~\ref{sec:design}).  
\item We implement a prototype of \algname based on the mainstream vLLM framework and experimentally demonstrate that, under heterogeneous workloads with mixed computing and I/O characteristics, \algname reduces the average job completion time (JCT) by up to 25.5\% compared to baseline methods, significantly improving the quality of experience (QoE) for latency-sensitive interactive applications . Additionally, it exhibits superior robustness, with its performance degrading 3.2x less than baselines as system load increases (Sec.~\ref{sec:evaluation}). 
\end{itemize}
\section{Background and Motivation}
\label{sec:background}
The rise of LLMs as autonomous agents represents a paradigm shift in computational workloads, moving from one-shot inference to dynamic, multi-stage workflows that interleave computation with external tool use. This new paradigm exposes critical bottlenecks in existing LLM serving systems, which are architected for one-shot queries. 
This section establishes the necessary background by contextualizing this evolution, characterizing the inference workflow, and demonstrating through concrete examples why a new scheduling-centric approach is essential for agentic efficiency.

\subsection{Background}

\subsubsection{ \textbf{Evolution from Basic LLMs to LLM-Powered Agents}}
Early LLMs excelled in in-context learning (ICL) but essentially operated as standalone text generators, lacking mechanisms for interacting with external environments or dynamically verifying generated knowledge. 
A pivotal advancement was the emergence of structured reasoning frameworks, such as Chain-of-Thought(CoT) \cite{ref20}. This transformed the inference workload from a single query\-response transaction into a longer, sequential reasoning process, implicitly introducing early workflow semantics.

Building upon this, frameworks like ReAct~\cite{ref50} formally integrated reasoning with external tool use, establishing the core agentic loop of interleaved local computation and remote I/O. This shift in capability corresponds directly to a shift in system demands: the workload is no longer a monolithic computation, but a dynamic graph of dependent segments with heterogeneous resource requirements (compute, memory, I/O).

These agentic capabilities are typically realized through various modes of API calls, which can be categorized into three primary types: integrating non-LLM tools (e.g., translators, retrieval systems), iterative self-calls for complex problem-solving, and multi-component collaboration (e.g., LLMs orchestrating other models). This paradigm has enabled sophisticated frameworks like Lang\-Chain~\cite{ref21} and AgentGraph~\cite{ref25}, which support complex, multi-agent workflows and present unprecedented challenges for resource management and scheduling in the serving infrastructure.


Based on this, frameworks such as ReAct~\cite{ref50} integrate reasoning with external actions, allowing models to leverage tools and ground their internal reasoning in real-world observations. From a systems perspective, this integration transforms the workload from continuous computation into a loop between local inference and remote I/O, which characterizes agentic workflows.

The ``action'' capability of agents is realized through various modes, which can broadly be seen as API calls. Depending on the interaction targets, they can be categorized into three main types: \textbf{(1) Integrating non-LLM tools}, where external dedicated APIs are called to compensate for an LLM's shortcomings, such as using a translator or an information retrieval system; \textbf{(2) Iterative self-calls}, where the LLM calls itself multiple times to decompose tasks and maintain a reasoning history; and \textbf{(3) Multi-component collaboration}, which combines multiple models and tools into a collaborative system orchestrated by a core planner LLM.


This paradigm gave rise to frameworks such as LangChain~\cite{ref21} and AgentGraph~\cite{ref25}, thereby enabling complex multi-agent collaborative workflows.

\subsubsection{\textbf{LLM Inference Workflow}}
LLMs drive various chatbot and AI applications, primarily utilizing Transformer-based models such as GPT~\cite{ref51}, Claude~\cite{ref52}, and LLaMA~\cite{ref53}. The LLM inference process generally consists of two main stages: Prefill and Decode. 

In the prefill stage, the system processes the input prompt, converting it into an intermediate token state through a single forward pass. This stage is computation-intensive as it requires processing the entire input to generate the initial KV cache, which stores intermediate state information needed for the model’s generation process, speeding up subsequent token generation.

Next, in the decode stage, the model generates new tokens one by one in an autoregressive manner, relying on previously generated tokens. This process is memory-bandwidth intensive, as the model frequently accesses the previously generated tokens and KV cache to generate the next tokens. Through this mechanism, the model can generate coherent outputs without recalculating all previous tokens, significantly improving efficiency.

However, in the case of LLM agent inference, the system may need to make external tool calls (e.g., querying a database, performing complex calculations, or accessing external APIs). These external calls interrupt the inference process, causing the system to enter a waiting state until the external tool returns its result. API call times are highly unpredictable, ranging from sub-millisecond API calls to human responses taking several minutes. 
At this point, the system must choose how to manage the interrupted KV cache sequence. Common strategies involve a trade-off between latency and memory overhead. The \textbf{Preserve} strategy keeps the KV cache fully in GPU memory, which improves resumption latency but increases memory pressure and reduces system throughput. At the other extreme, the \textbf{Discard} strategy releases all KV cache and re-executes the prefill stage once the tool returns; this simplifies state management but leads to redundant computation. A hybrid approach is the \textbf{Swap} strategy, which migrates the KV cache to CPU memory during the interruption and moves it back afterward, balancing memory pressure and computational efficiency at the cost of I/O transfer delays.

After the API call ends, its response is treated as an additional token sequence, appended to the original input sequence. The model only needs to perform a forward pass on the newly added part and append the generated key-value cache to the original cache.

\subsection{Motivation}

Although often designed to minimize end-to-end latency, traditional schedulers prove counterproductive in agentic workflows. By treating each computational segment as an independent task, they adopt a myopic perspective.
While this may optimize local performance, it neglects the chained dependencies inherent in a request's lifecycle, leading to suboptimal global performance and a decrease in overall system efficiency.
\begin{figure}[htbp]
    \centering
    \includegraphics[width=0.9\linewidth]{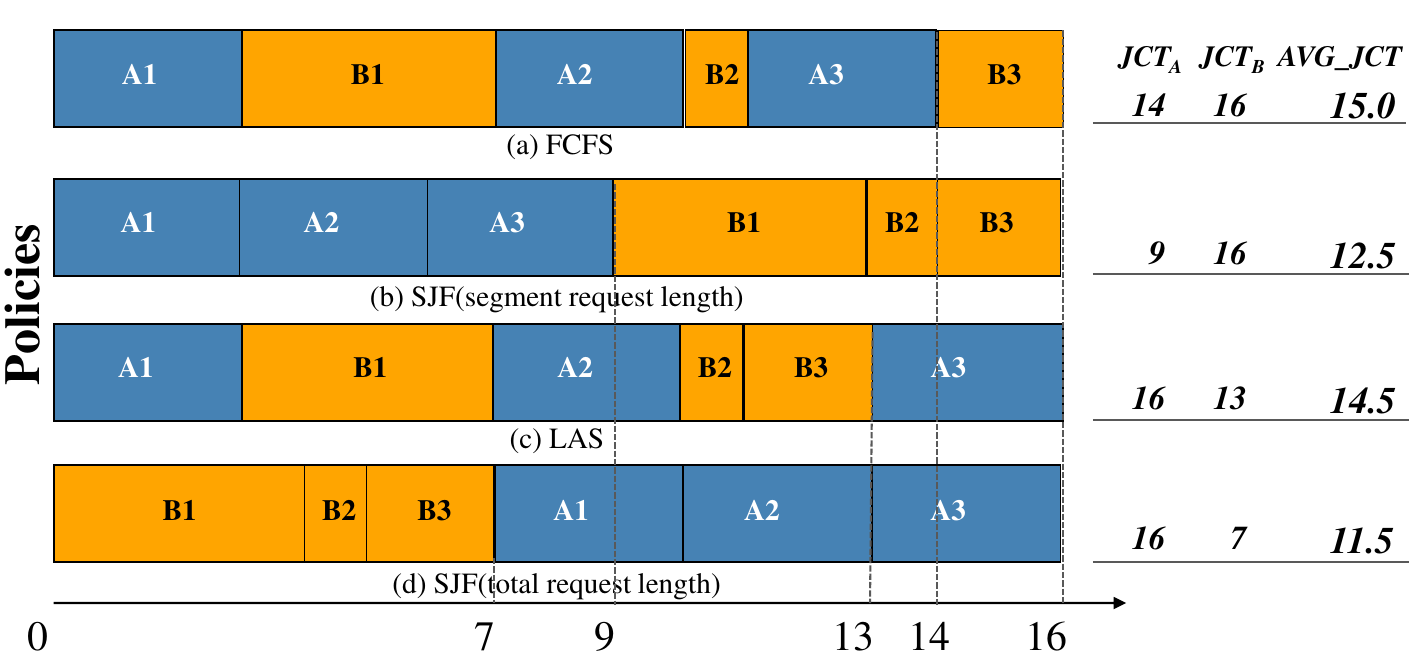}
    \caption{Gantt charts visualizing the execution timelines for the two requests under the four scheduling policies. }
    \label{fig:motivation_gantt}
    \Description{This is an example image showing ...}
\end{figure}

To illustrate this limitation, we present a motivating example with two concurrent requests, A and B, whose segment execution times are detailed in Figure~\ref{fig:motivation_gantt}. Request A consists of three segments, each requiring 3 time units to process. Request B is composed of three segments with processing times of 4, 1, and 2 time units, respectively. We compare the performance of four scheduling policies in this scenario:
\begin{itemize}
    \item \textbf{First-Come-First-Served (FCFS):} A baseline policy that executes segments strictly by their arrival order. In this example, its arbitrary initial choice (prioritizing A1 over B1) creates a cascading delay for B's subsequent segments due to the chained dependencies of agentic workflow, resulting in a poor average JCT of 15.0.
    \item \textbf{Shortest-Job-First (SJF) at the segment level:} A myopic policy that always selects the ready segment with the shortest processing time. While this strategy rapidly completes all of Request A's short segments, achieving an excellent local JCT of 9.0 for A, it completely starves the longer Request B until A finishes, resulting in a bad JCT of 16.0 for B and a mediocre average JCT of 12.5.
    \item \textbf{Least-Attained-Service (LAS):} A fairness-oriented policy that prioritizes the request which has received the least cumulative service. It interleaves A and B, but fails to recognize that prioritizing B's short final segments (B2, B3) would be more globally efficient, resulting in a high average JCT of 14.5.
    \item \textbf{Shortest-Job-First (SJF) at the request level:} An oracle policy that prioritizes the request with the shortest total service time, representing the optimum. It intelligently prioritizes the completion of the shorter Request B first, thereby minimizing the average JCT to 11.5.
\end{itemize}

The analysis reveals that while traditional scheduling policies possess distinct merits, such as the simplicity and starvation-free guarantee of FCFS, the inter-request fairness of LAS and the proven effectiveness of segment-level SJF's greedy optimization in one-shot inference, their shared inability to model the multi-stage leads to severe cascading delays or HoL blocking. In this work, we propose \algname, a state-aware scheduling engine designed to transcend these limitations. By unifying historical context with predictive forecasting, \algname approximates the benefits of a global, lifecycle-oriented perspective in an online setting.

\section{System Model and Problem Formalization}
\label{sec:system_model}

To address the scheduling challenges of alternating ``compute-I/O'' execution in LLM agent inference, this section will first formally models the agentic execution workflow and then defines the scheduling objectives.

\subsection{Problem Formalization}
\label{sec:problem_formalization}
LLM agents accomplish complex tasks through using external tools, making their execution inherently multi-stage.
We define an end-to-end inference task initiated by a user as a \textit{parent request} (or complete request), denoted as \( R_i \). Its execution process is not a single computational process but a dynamic, multi-stage workflow composed of several segments. We formally represent the lifecycle of \( R_i \) as an ordered sequence of segmented requests:
\[ R_i = \langle S_i^1, S_i^2, \dots, S_i^{k} \rangle,\]
where \( k_i \) is the total number of segments in request \( R_i \), which is unknown upon the request's initial arrival.

The internal execution flow of each segmented request \( S_i^j \) can be further abstracted into three ordered stages with distinct resource consumption characteristics:
\begin{itemize}
    \item \textbf{Prefill Stage:} This stage is responsible for a parallel Attention computation on the current context to generate the initial KV cache. Its execution speed is primarily limited by the GPU's floating-point operations per second (FLOPS). Therefore, we classify this stage as \textbf{Compute-bound}.
    \item \textbf{Decoding Stage:} This stage generates tokens one by one in an autoregressive manner. Its execution speed is mainly limited by the GPU's memory bandwidth. We classify this stage as \textbf{Memory-bound}.
    \item \textbf{API Call Stage:} This stage involves interaction with external network services, during which local compute and memory resources remain idle. We classify this stage as \textbf{I/O-bound}.
\end{itemize}

To formally model the segmented request \( S_i^j \), we represent it as a triplet, where each element represents the expected service time cost of the corresponding stage:
\[ S_i^j = (T(P_i^j), T(D_i^j), T(A_i^j)). \]
The alternating presence of different resource bottlenecks within a request's lifecycle is the core challenge of this scheduling problem. Traditional schedulers typically treat each ready computational task (i.e., segment \( S_i^j \)) as an independent, stateless scheduling unit, which systematically penalizes requests with multiple I/O interruptions.

The core scheduling objective of this research is defined as minimizing the average end-to-end Job Completion Time (JCT) for all requests. We define the JCT of a complete request \( R_i \) as the total duration from its arrival time, \( t_{\text{arrival}}(R_i) \), to the completion of its final segment's computation, \( t_{\text{finish}}(R_i) \). 
This duration is composed of the actual processing time of all its segments and the waiting time in the ready queue. Let \( C(S_i^j) \) be the completion time of segment \( S_i^j \), then \( t_{\text{finish}}(R_i) = C(S_i^{k}) \). The completion time of a segment can be recursively defined as:
\begin{equation}
\label{eq:completion_time}
C(S_i^j) = 
\begin{cases} 
      t_{\text{arrival}}(R_i) + W(S_i^1) + T_{\text{comp}}(S_i^1) & \text{if } j=1 \\
      C(S_i^{j-1}) + T(A_i^{j-1}) + W(S_i^j) + T_{\text{comp}}(S_i^j) & \text{if } j>1 
\end{cases},
\end{equation}
where \( T_{\text{comp}}(S_i^j) \) and \( T(A_i^j) \) represent the actual time spent in the computation stages (prefill and decoding) and the API call stage of segment \( S_i^j \), respectively. \( W(S_i^j) \) is the waiting time of the segment in the ready queue before being scheduled for execution. 
Thus, the JCT of request \( R_i \) can be expressed as:
\begin{equation}
\label{eq:jct}
JCT(R_i) = C(S_i^k) - t_{\text{arrival}}(R_i) = \sum_{j=1}^{k} \left( W(S_i^j) + T_{\text{comp}}(S_i^j) \right) + \sum_{j=1}^{k-1} T(A_i^j),
\end{equation}
the problem's optimization objective can be formally defined as:
\begin{equation}
\label{eq:objective}
\text{Minimize} \quad \frac{1}{|\mathcal{R}|} \sum_{R_i \in \mathcal{R}} \text{JCT}(R_i),
\end{equation}
where \( \mathcal{R} \) represents the set of all requests processed by the system.

\subsection{Problem Complexity}
\label{sec:np_proof}

We prove that the scheduling problem defined in Section~\ref{sec:problem_formalization} is NP-hard.

\textbf{Theorem 3.1.} \textit{The LLM agent scheduling problem is NP-hard.}

\begin{proof}
The detailed proof by reduction from the classic\\ $1|prec|\sum C_j$ problem is provided in Appendix~\ref{app:np_proof_detail}.
\end{proof}

\section{Scheduling Mechanism Design}
\label{sec:design}

\subsection{Overall Architecture}

\begin{figure}[!ht]
    \centering
    \includegraphics[width=0.95\linewidth]{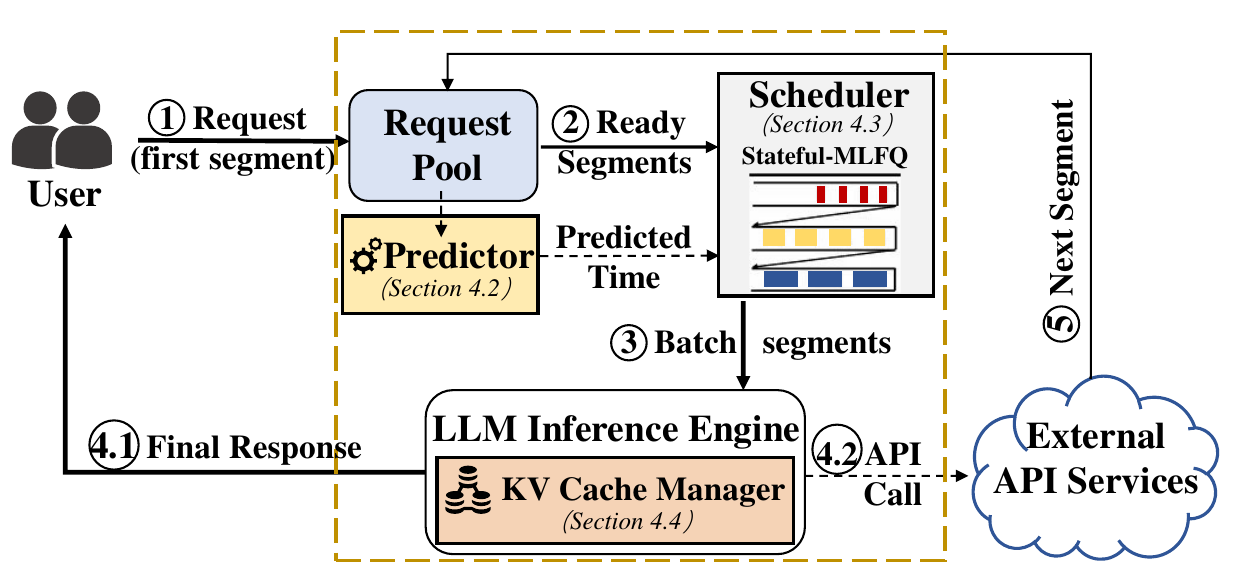} 
    \caption{The overall architecture of \algname.}
    \label{fig:architecture}
    \Description{This is an example image showing ...}
\end{figure}
To address the complexities of agentic workflows, we have designed a scheduling system named \algname. Its architecture, illustrated in Figure~\ref{fig:architecture}, is composed of several key components designed to work in concert: a unified Request Pool, a \textbf{Service Time Predictor}, a state-aware \textbf{Scheduler}, an LLM Inference Engine, and an adaptive \textbf{KV Cache Manager}.The three core components are overviewed below to illustrate their primary functions.
\begin{itemize}
\item \textbf{Service Time Predictor (Sec.~\ref{sec:predictor_design}):} This component annotates request segments with estimated computation time and API duration. It combines offline profiling for prefill latency, a segment-level generation length oracle, and category-based API latency statistics to provide essential metadata for scheduling decisions.
\item \textbf{Stateful-MLFQ Scheduler (Sec.~\ref{sec:scheduling_algorithm}):}  As the system's core decision-maker, this scheduler implements a multi-level feedback queue algorithm that dynamically classifies requests based on compute and I/O behavior. It uses token cost thresholds for priority migration and an enhanced HRRN policy for intra-queue ordering to balance efficiency and fairness.
\item \textbf{KV Cache Manager (Sec.~\ref{sec:kv_cache_management}):} This manager handles the latency-throughput tradeoff during I/O waits by dynamically selecting strategy based on GPU memory pressure to minimizes memory waste.
\end{itemize}
These components operate in a coordinated, cyclic workflow: After ready segments are collected in the Request Pool, they are annotated by the Predictor with crucial metadata, including estimated computation times and API call durations. The Scheduler then prioritizes and batches them for execution by the Inference Engine. Once a segment triggers an external API call, its state is managed by the KV Cache Manager during the wait. This orchestrated workflow enables global optimization across the entire request lifecycle.


\subsection{Service Time Predictor Design}
\label{sec:predictor_design}
The efficacy of our scheduling decisions hinges on accurate service time predictions for each stage of a request. The design of the predictor and the scheduling algorithm are orthogonal problems. To rigorously evaluate our scheduler's performance, isolated from potential prediction inaccuracies, we employ an idealized yet practical prediction methodology.

\paragraph{Compute Time Prediction.} The total computation time, \( T_{\text{comp}}(S_i^j) \), is composed of prefill and decoding time. For the prefill stage, which is a deterministic function of input length, we build a data-driven performance model by offline-profiling the target hardware with various sequence lengths. For the decoding stage, predicting the generation length is a known challenge. Prior work has demonstrated high-accuracy predictors are feasible. Based on this, we adopt a \textbf{segment-level oracle} for the number of generated tokens ($n_\text{gen}$), using the ground-truth value from our dataset. It is critical to note this oracle has no knowledge of future segments, thus preserving the online nature of the scheduling problem. The total predicted computation time is then: \(T_{\text{comp}}(S_{i}^{j}) = f_{\text{prefill}}(n_{\text{in}}) + n_{\text{gen}} \cdot \text{avg\_decode\_latency\_per\_token}\).

\paragraph{API Latency Prediction.} API call durations are highly variable and depend on external factors. Static analysis is often intractable. However, we observe that APIs within the same functional category exhibit stable latency distributions. For example, math-related API calls average 9e-5 seconds, while image generation and chatbot APIs can take tens of seconds (e.g., means of 20.03s and 28.6s respectively). Leveraging this, we build a statistical model based on these categories. During scheduling, we extract the API category from the prompt and use the category's mean latency as the predicted value \(T_A(S_{i}^{j})\).

\subsection{Stateful-MLFQ Scheduling Algorithm}
\label{sec:scheduling_algorithm}
\subsubsection{Algorithm Design Overview}
To operationalize the global optimization objective defined in Section~\ref{sec:problem_formalization}, we design a state-aware multi-level feedback queue scheduling algorithm (Stateful-MLFQ), with its full logic presented in Algorithm~\ref{alg:stateful_mlfq}. The core of this algorithm lies in its ``Stateful'' nature: it not only evaluates a segment's current characteristics (e.g., estimated service time) but also unifies the parent request's historical behavior (e.g., accumulated wait time, past compute and I/O patterns) with future predictions into a single decision-making framework. It is a hierarchical, preemptive scheduling algorithm that aims to strike a balance between efficiency and fairness through macro and micro-level controls.

\subsubsection{Macro-level Control: Event-driven Priority Migration}
The algorithm's macro-level framework is built upon a multi-level feedback queue (MLFQ) structure, consisting of $m$ queues $Q_0, ..., Q_{m-1}$ with strict priorities.

We adopt Token Cost instead of a time slice as the migration threshold because the time slice is an unstable metric heavily influenced by batch composition in continuous batching. In contrast, Token Cost is a deterministic, intrinsic metric that solely measures the computational work a request has received. This stability makes priority migration decisions fairer and more robust.

The migration of requests between queues is event-driven, as defined in the event-handling functions of Algorithm~\ref{alg:stateful_mlfq}.
\begin{enumerate}
    \item \textbf{On Request Arrival:} All new requests are placed into the highest-priority queue $Q_0$ to ensure a fast response.
    \item \textbf{On Segment Completion:} After a segment finishes execution, the system dynamically adjusts its parent request's priority based on its behavior.
    \begin{itemize}
        \item \textbf{Demotion:} If a segment's computational cost exceeds its queue's threshold, the parent request is identified as compute-intensive and demoted to the next lower-priority queue.
        \item \textbf{Promotion:} If a segment yields to an API call before exhausting its token cost quota, the parent request is identified as I/O-intensive and is promoted to a higher-priority queue. This policy aims to prioritize I/O-bound requests to minimize their impact on the total JCT.
    \end{itemize}
\end{enumerate}

\begin{algorithm}[htbp]
    \SetKwInOut{Input}{Input}
    \SetKwInOut{Output}{Output}
    \Input{
        $m$: Number of queues; $Q$: Priority queues; $T$: Token thresholds; $\tau$: Aging threshold.
    }
    \Output{
        The next execution batch $B$.
    }
    \BlankLine
    \SetKwFunction{FBuildNextBatch}{BuildNextBatch}
    \SetKwProg{Fn}{function}{:}{}
    \Fn{\FBuildNextBatch{$Q, T, \tau$}}{
        \tcp{1. Starvation Prevention (Aging)}
        \ForEach{request $R \in Q_{m-1}$}{
            \If{$(R.waitTime + R.nextprocTime) / R.nextprocTime > \tau$}{
                $Q_0 \leftarrow Q_0 \cup \{R\}$\;
                $Q_{m-1} \leftarrow Q_{m-1} \setminus \{R\}$\;
            }
        }
        \BlankLine
        \tcp{2. Batch Construction}
        $B \leftarrow \emptyset$\;
        \For{$k \leftarrow 0$ \KwTo $m-1$}{
            \If{$Q_k$ is not empty}{
                $C \leftarrow \text{GetAllReadySegments}(Q_k)$\;
                \tcp{3. Intra-queue sorting using HRRN}
                \ForEach{segment $S \in C$}{
                     $S.HRRNscore \leftarrow (S.totalwaitTime + S.procTime) / S.procTime$\;
                }
                $C_{sorted} \leftarrow \text{SortByHRRNScore}(C, \text{DESC})$\;
                \tcp{4. Pack batch respecting memory constraints}
                \ForEach{segment $S \in C_{sorted}$}{
                     \If{CanFitInMemory($B \cup \{S\}$)}{
                        $B \leftarrow B \cup \{S\}$\;
                    }
                }
                \KwRet{$B$}\;
            }
        }
        \KwRet{$B$}\;
    }
    \caption{Stateful-MLFQ Scheduling}
    \label{alg:stateful_mlfq}
\end{algorithm}

\subsubsection{Scheduling Cycle: Batch Building and Intra-Queue Sorting}
In each scheduling cycle, our core scheduling function, \texttt{BuildNextBatch}, is invoked to determine the next batch of requests to execute.

The function's first step is to handle starvation (lines 1-6). It inspects all requests in the lowest-priority queue, $Q_{m-1}$, and if a request's response ratio exceeds a predefined aging threshold, it is preemptively promoted to the highest-priority queue, $Q_0$.

After handling starvation, the scheduler iterates from the highest-priority queue $Q_0$ downwards (line 8) to find the first non-empty queue, $Q_k$. For all ready segments within this queue, the scheduler performs micro-level, intra-queue sorting (lines 11-14). We employ the Highest Response Ratio Next (HRRN) policy to calculate a score for each segment:
\begin{equation}
Score_{HRRN}(S_{i}^{j})=\frac{W(R_{i})+T_{\text{proc}}(S_{i}^{j})}{T_{\text{proc}}(S_{i}^{j})},
\end{equation}
where $W(R_i)$ is the accumulated waiting time of its parent request and $T_{\text{proc}}(S_{i}^{j})$ is the estimated service time of the current segment. This mechanism behaves like Shortest Remaining Processing Time (SRPT) when waiting times are comparable, enhancing efficiency. As the waiting time of a long job accumulates, its score increases, ensuring intra-queue fairness.

Finally, after sorting candidate segments by their HRRN score, the scheduler packs them sequentially into the next batch until GPU memory capacity is reached (lines 15-20).

\subsubsection{Preemption Granularity}
It is important to note that our algorithm's preemption occurs at the segment level. Once a batch of tasks begins execution, it runs to completion (i.e., until all segments in the batch either trigger an API call or generate a final response) without being interrupted at the iteration level. Preemption is realized in each new scheduling cycle: a newly arrived or promoted high-priority request can ``preempt'' the execution opportunity of a lower-priority request when the next batch is being constructed. This design avoids the prohibitive overhead of fine-grained preemption and its associated KV cache swapping costs.
\subsection{KV Cache Management}
\label{sec:kv_cache_management}
The preemptive nature of our scheduling mechanism necessitates the preservation of intermediate states (KV cache) for all preempted yet incomplete requests.
Without an effective management policy, GPU memory would become a critical bottleneck, limiting the scheduler's efficacy and potentially reintroducing the HoL blocking we aim to solve. 
To address this, \algname integrates an adaptive KV cache management policy designed to dynamically balance single-request latency with overall system throughput. The policy is governed by a high-watermark threshold for GPU memory usage, creating two distinct operational modes. In low-load scenarios (i.e., below the threshold), the system defaults to a \textbf{Preserve} policy for all I/O-bound requests to prioritize low latency.

Conversely, when memory pressure is high, the objective shifts to maximizing resource utilization by minimizing memory-time waste. The system evaluates the potential waste for three candidate policies: \textbf{Preserve}, \textbf{Discard}, and \textbf{Swap}, choosing the one with the minimum cost. Following the model proposed by Infercept~[1], the waste ($W$) for each policy is estimated as:
\begin{align}
    W_{\text{preserve}} &= T_{\text{api}} \cdot C_{\text{self}} \cdot M, \\
    W_{\text{discard}} &= T_{\text{recompute}} \cdot C_{\text{batch}} \cdot M, \\
    W_{\text{swap}} &= 2 \cdot T_{\text{swap}} \cdot C_{\text{batch}} \cdot M.
\end{align}
Here, $T_{\text{api}}$, $T_{recompute}$, and $T_{swap}$ are the predicted durations for the API call, KV cache recomputation, and swap I/O, respectively. $C_{self}$ is the token count of the request's own cache, $C_{batch}$ is the total token count of other requests that could be batched if memory were freed, and $M$ is the memory required per token's KV cache.

The system then selects the optimal strategy that results in the least memory waste:
\begin{equation}
\text{Strategy} = \mathop{\arg\min}\limits_{s \in \{\text{Preserve, Discard, Swap}\}} W_s.
\end{equation}
This adaptive policy ensures high resource utilization under contentious loads while maintaining responsiveness in uncongested scenarios.

\section{Experimental Evaluation}
\label{sec:evaluation}

\subsection{Experimental Setup}

\paragraph{Testbed Implementation} We implemented the \algname system on top of vLLM, retaining its high-efficiency inference architecture while improving upon its original KV Cache management. \algname introduces a dynamic scheduling policy based on GPU memory utilization. When memory availability is sufficient, the system prioritizes reducing request latency by enforcing the ``Preserve'' policy to ensure a fast response. Conversely, when memory resources become scarce, the system automatically adjusts its policy to prioritize maximizing overall resource utilization, employing the ``Discard'' and ``Swap'' strategies to balance memory occupation and system throughput. Our improvements in memory management and scheduling are highly modular, allowing for seamless integration with other non-interfering LLM optimization methods.

\paragraph{Environment} The experiments were conducted on a machine equipped with two NVIDIA A100 80GB GPUs connected via NVLink, a 28-core Intel Xeon Gold 6330 CPU at 2.0GHz, and 503GB of RAM. To evaluate the generality and robustness of the proposed algorithm across different model scales, we conducted experiments on two open-source large models: the 6B-parameter GPT-J~\cite{ref41} and Vicuna-13B~\cite{ref42}. GPT-J was run on a single A100, suitable for \\medium inference load scenarios, while Vicuna-13B was run in a multi-GPU environment with two A100s to create higher memory pressure scenarios, testing the policy's performance under complex resource contention conditions.

\paragraph{Dataset} We used the dataset released by Infercept for our experiments. This dataset comprises six sub-tasks:
\begin{itemize}
    \item \textbf{Arithmetic Operations:} Based on the GSM8K-XL dataset \cite{ref43}, covering math problems that require multi-step reasoning.
    \item \textbf{Knowledge Question Answering:} Based on the MultiHop QA dataset~\cite{ref44}, representing knowledge base retrieval requests.
    \item \textbf{Virtual Environment:} Based on the ALFWorld dataset~\cite{ref45}, simulating an LLM controlling entities in a virtual environment to complete tasks.
    \item \textbf{Multi-turn Dialogue:} Based on the ShareGPT dataset~\cite{ref46}, simulating multi-turn human interactions.
    \item \textbf{Image Generation:} Using ChatGPT~\cite{ref47} to automatically generate prompt sequences that trigger calls to the Stable Diffusion model~\cite{ref48}.
    \item \textbf{Text-to-Speech:} Using ChatGPT to generate prompt sequences that trigger calls to the Bark TTS model~\cite{ref49}.
\end{itemize}
The original dataset was uniformly sampled from these six task types. However, to better highlight the performance differences among various scheduling and memory policies, we artificially increased the sampling proportion of long-latency API requests in our experiments to construct a more challenging system load.

\paragraph{Baseline Methods} Our experiments will compare \algname with the following four different policies: (1) vLLM with FCFS (First-Come-First-Served) scheduling; (2) vLLM with SJF (Shortest Job First) scheduling; (3) vLLM with LAS (Least Attained Service) scheduling, which is the core scheduling idea adopted by Autellix~\cite{ref36}; and (4) Infercept (with its default FCFS policy).

\paragraph{Metrics} Our primary evaluation metric is the average Job Completion Time (JCT), which measures the end-to-end latency from request submission to final completion. Unlike intermediate metrics like Time-To-First-Token (TTFT) or Time-Per-Output-Token (TPOT), JCT provides a holistic measure of system performance and user experience for multi-stage agentic workflows.

\subsection{End-to-End Performance Analysis}
\begin{figure}[h!]
    \centering 
    \captionsetup[subfigure]{skip=1pt}
    \begin{subfigure}[b]{0.49\columnwidth}
        
        \includegraphics[width=\columnwidth]{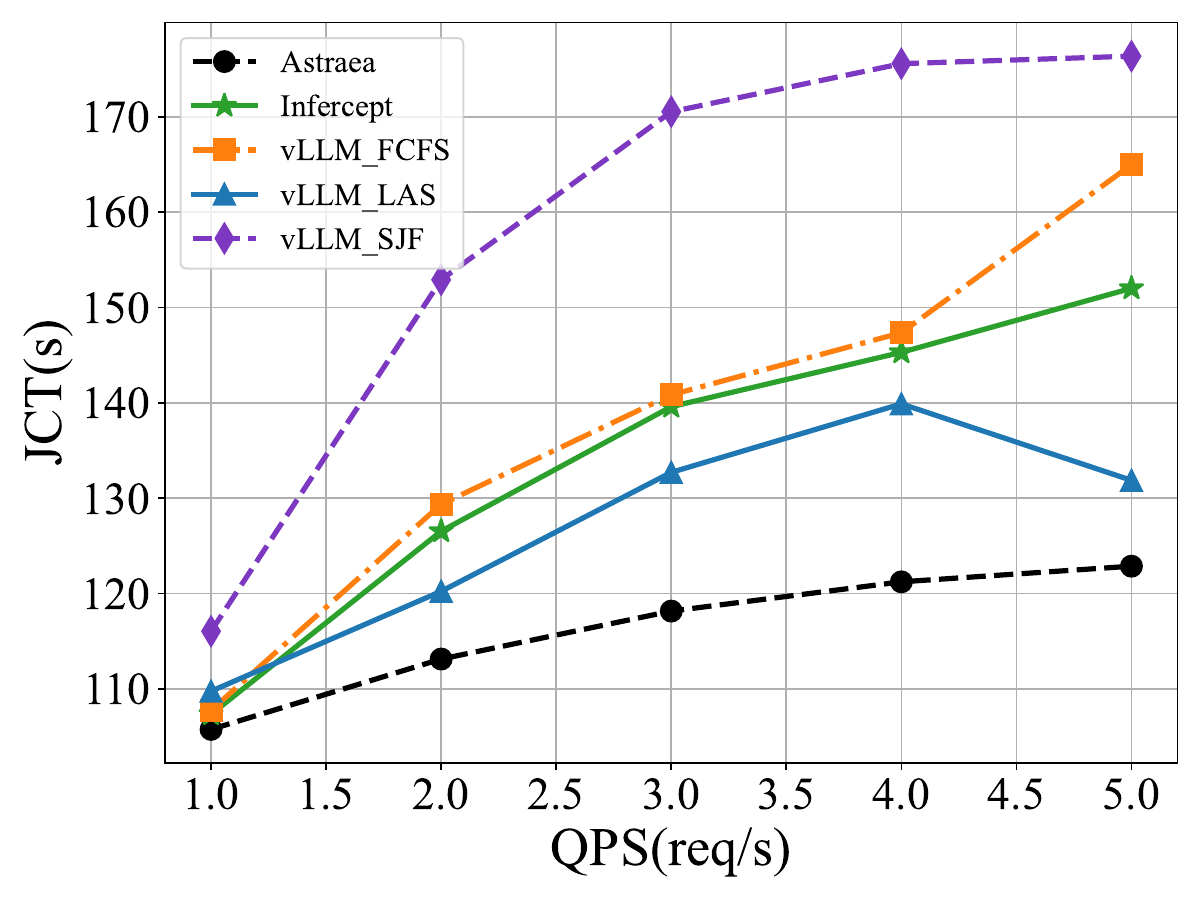} 
        \caption{30\% memory availability}
        \label{fig:gptj_30}
    \end{subfigure}
    \hfill 
    \begin{subfigure}[b]{0.49\columnwidth}
        
        \includegraphics[width=\columnwidth]{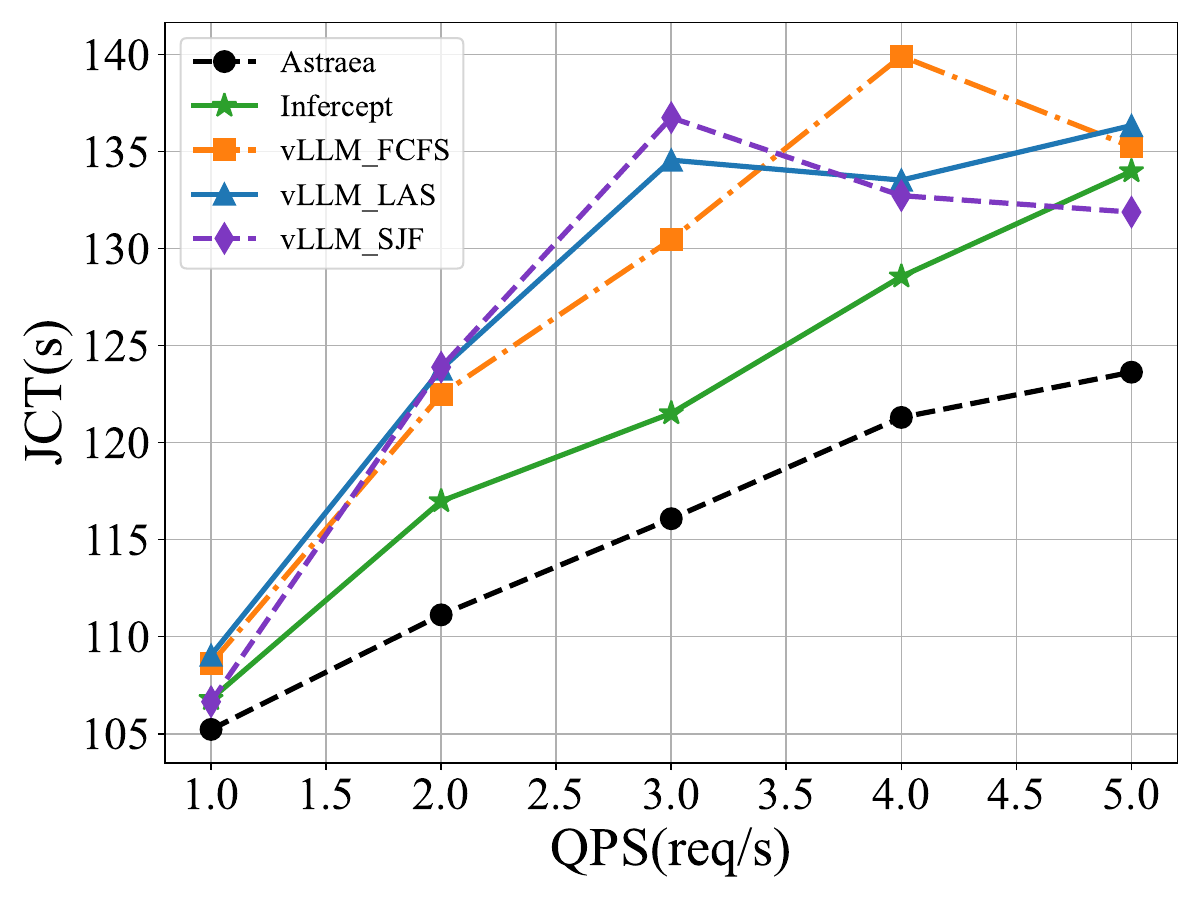} 
        \caption{50\% memory availability}
        \label{fig:gptj_50}
    \end{subfigure}
    \begin{subfigure}[b]{0.49\columnwidth}
        
        \includegraphics[width=\columnwidth]{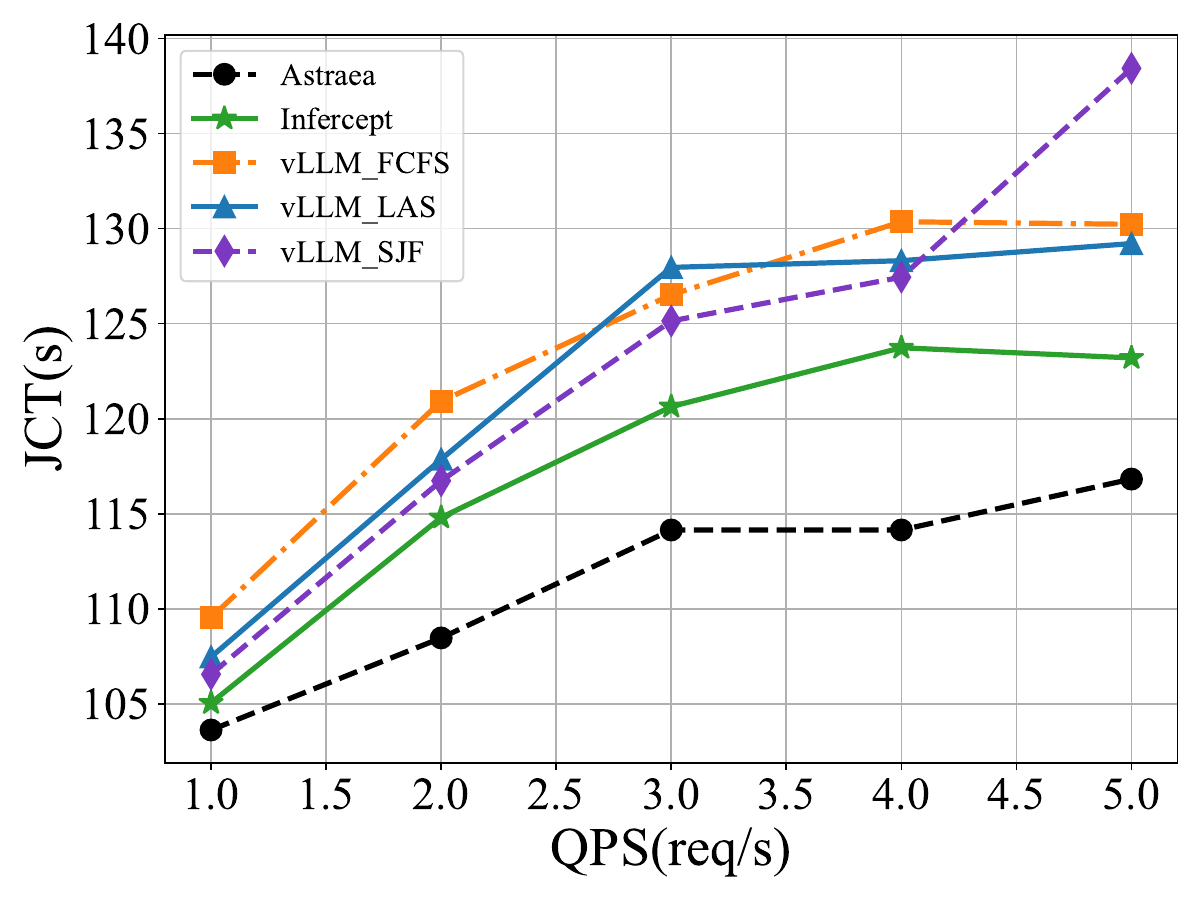} 
        \caption{70\% memory availability}
        \label{fig:gptj_70}
    \end{subfigure}
    \hfill
    \begin{subfigure}[b]{0.49\columnwidth}
        
        \includegraphics[width=\columnwidth]{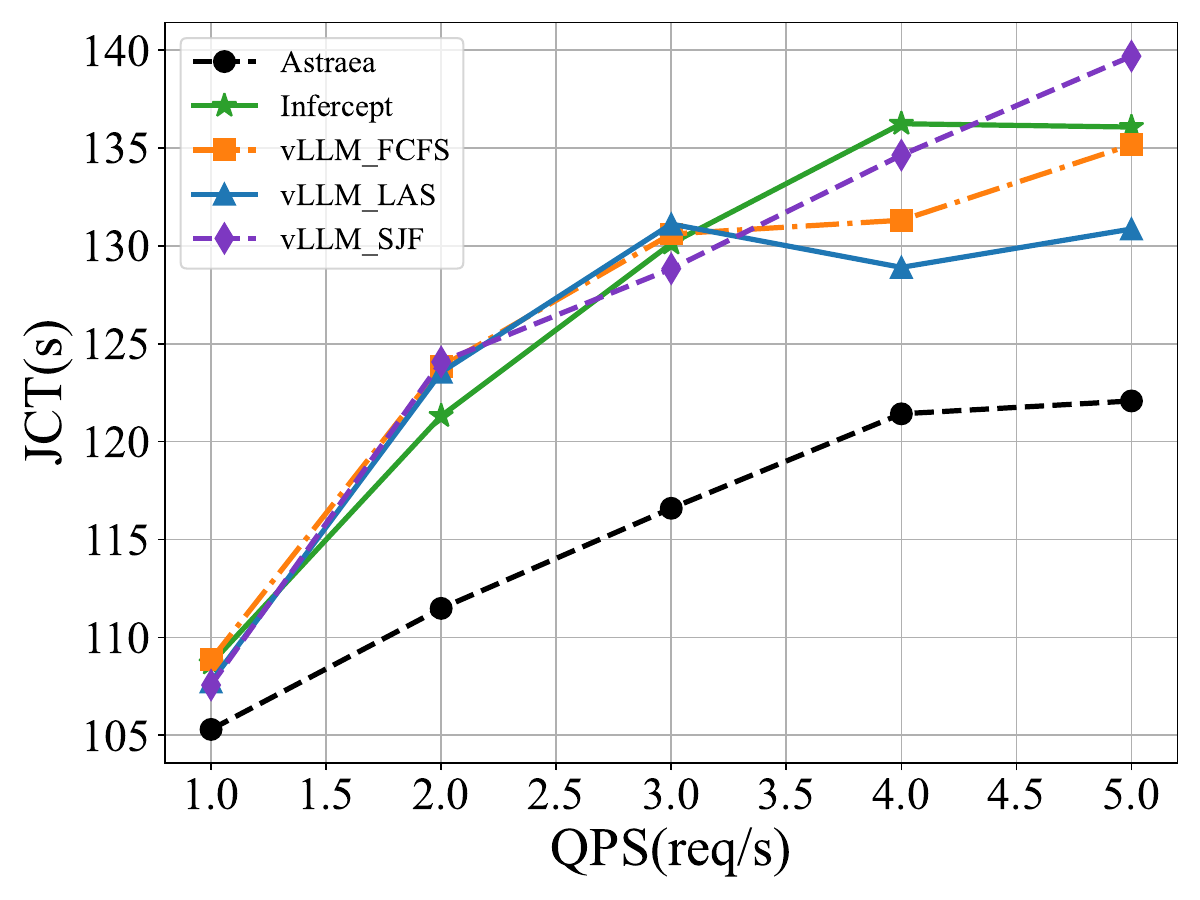} 
        \caption{90\% memory availability}
        \label{fig:gptj_90}
    \end{subfigure}

    \caption{Comparison of average JCT of various scheduling policies on the GPT-J model.}
    \label{fig:gptj_perf} 
    \Description{This is an example image showing ...}
\end{figure}

\begin{figure}[h!]
    \centering 
    \captionsetup[subfigure]{skip=1pt}
    \begin{subfigure}[b]{0.49\columnwidth}
        
        \includegraphics[width=\columnwidth]{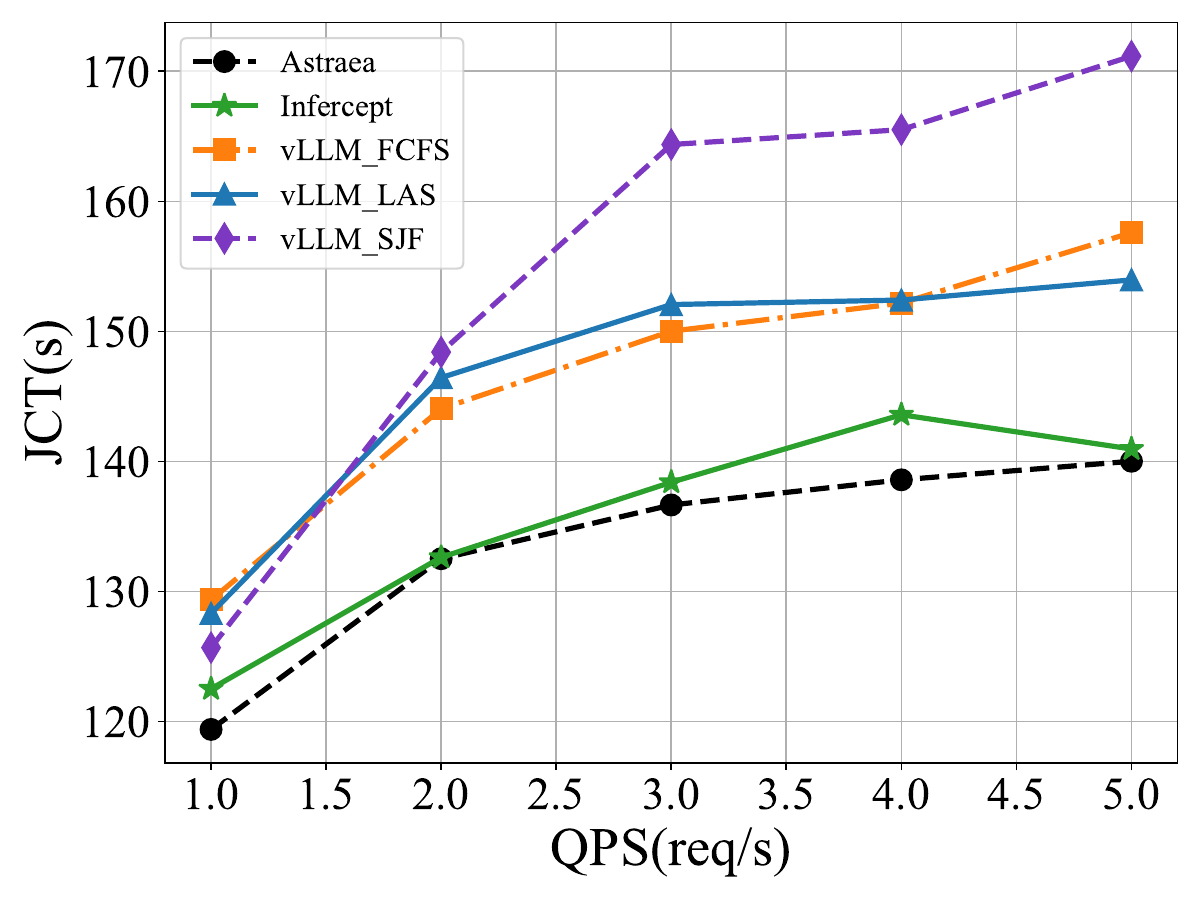} 
        \caption{30\% memory availability}
        \label{fig:vicuna_30}
    \end{subfigure}
    \hfill 
    \begin{subfigure}[b]{0.49\columnwidth}
        
        \includegraphics[width=\columnwidth]{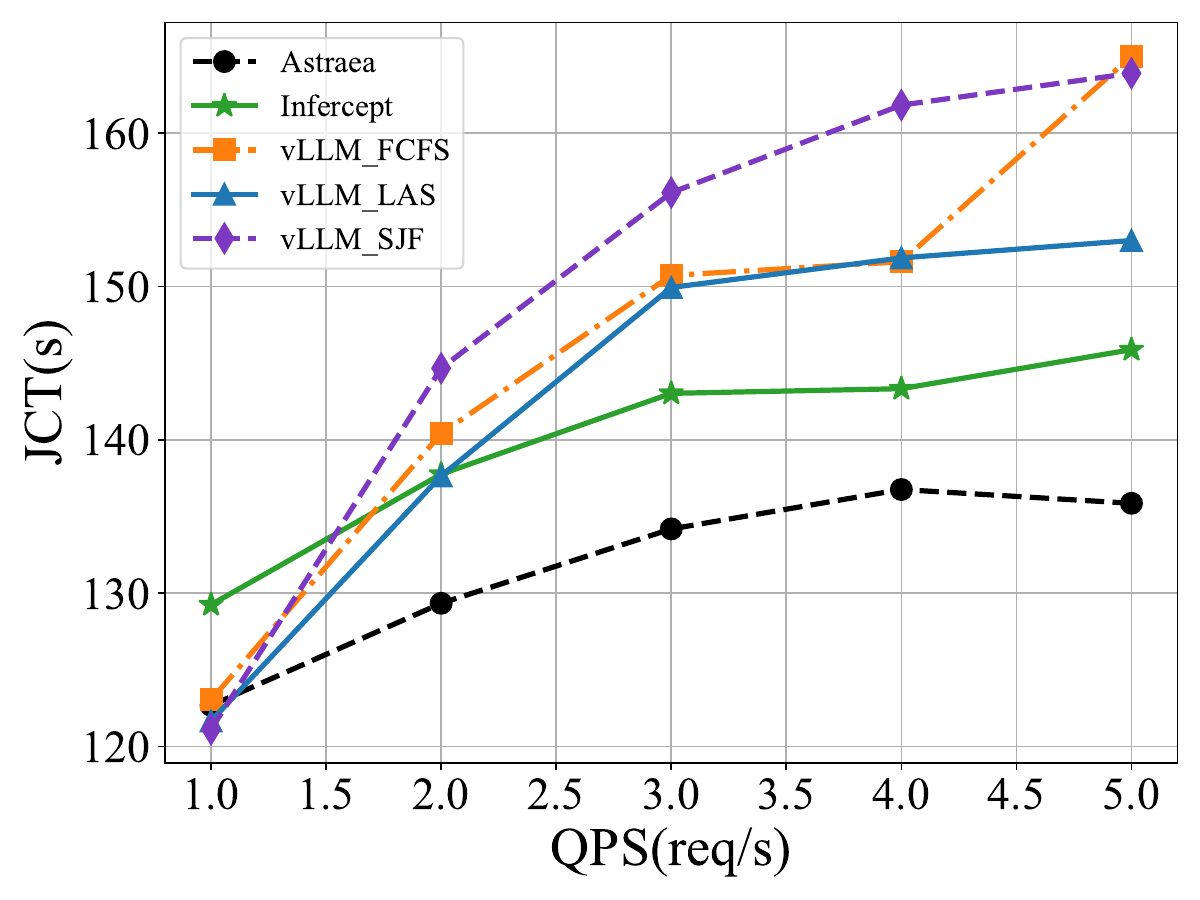} 
        \caption{50\% memory availability}
        \label{fig:vicuna_50}
    \end{subfigure}
     \begin{subfigure}[b]{0.49\columnwidth}
        
        \includegraphics[width=\columnwidth]{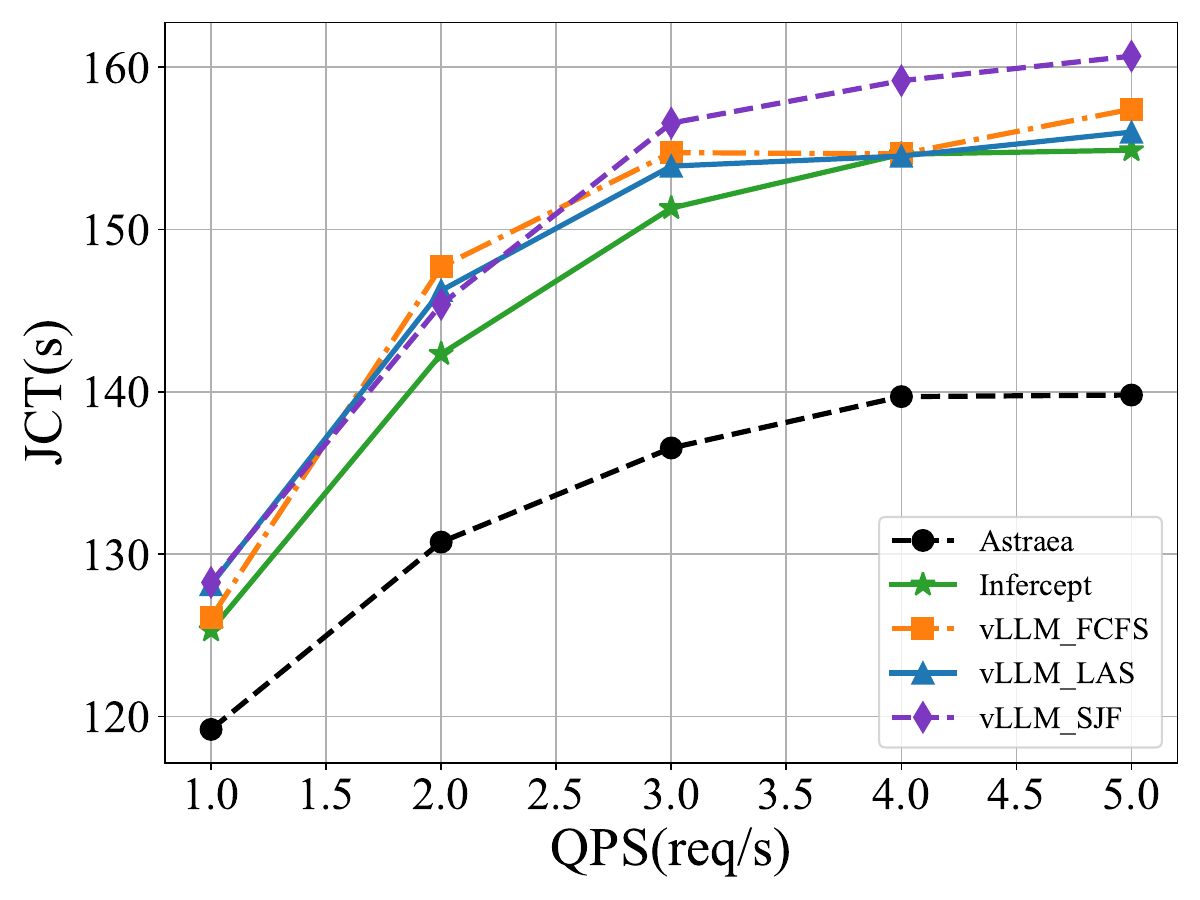} 
        \caption{70\% memory availability}
        \label{fig:vicuna_70}
    \end{subfigure}
    \hfill 
    \begin{subfigure}[b]{0.49\columnwidth}
        
        \includegraphics[width=\columnwidth]{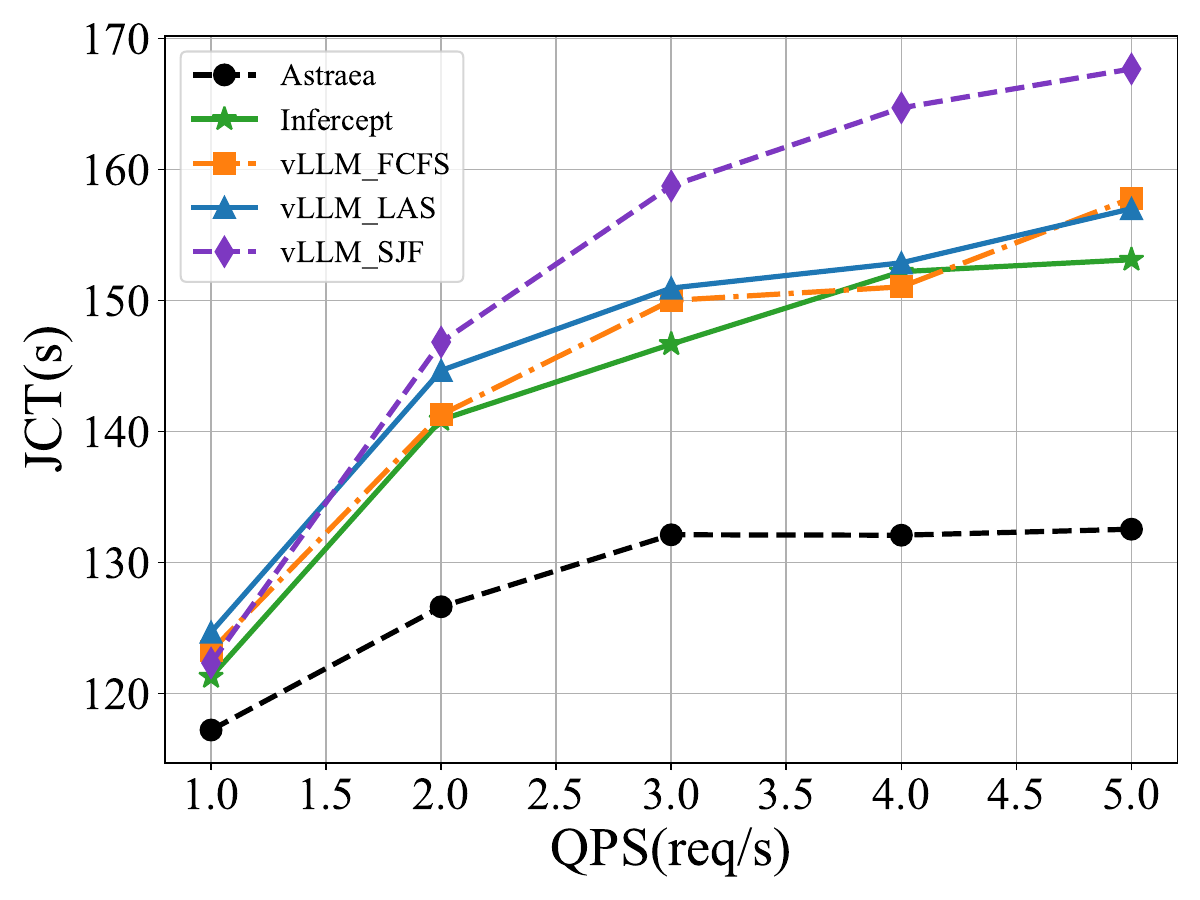} 
        \caption{90\% memory availability}
        \label{fig:vicuna_90}
    \end{subfigure}

    \caption{Comparison of average JCT of various scheduling policies on the Vicuna-13B model.}
    \label{fig:vicuna_perf} 
    \Description{This is an example image showing ...}
\end{figure}

We conducted a series of experiments on both the 6B-parameter GPT-J model and the larger 13B-parameter Vicuna model. To evaluate performance under varying system load and memory pressure, we measured the average JCT across a range of Queries Per Second (QPS) at four distinct GPU memory availability levels: 30\%, 50\%, 70\%, and 90\%. The results for both models are presented in Figure~\ref{fig:gptj_perf} and Figure~\ref{fig:vicuna_perf}.

On the GPT-J model (Figures~\ref{fig:gptj_perf}), \algname outperforms all baseline methods. The performance advantage is most pronounced under high memory pressure (30\% and 50\% availability). In the most contentious scenario with 30\% memory availability, the JCT of all baseline methods rises sharply with increasing QPS, indicating severe performance degradation. In contrast, \algname's performance curve remains flatter, demonstrating its superior stability. For example, at QPS=5, \algname's JCT is 122.88s, which is 19.1\% lower than Infercept and 25.5\% lower than vLLM-FCFS.

This performance advantage is even more critical on the larger Vicuna-13B model, where the higher memory footprint of the model itself intensifies resource contention. As shown in Figures~\ref{fig:vicuna_perf}, the performance degradation of baseline methods under memory pressure is more severe. At 30\% memory availability and a high load of QPS=5, \algname achieves a JCT of 140.02s. This is a significant improvement over vLLM-SJF (171.16s) and vLLM-LAS (153.96s), showcasing \algname's robustness when scheduling for larger models. Even in the high-availability 90\% memory scenario, \algname's JCT at QPS=5 (132.53s) is 13.4\% and 16.0\% lower than Infercept and vLLM-FCFS, respectively.

In summary, the results across both models validate our state-aware scheduling and adaptive cache management are effective. The benefits of \algname are particularly magnified in challenging, resource-constrained environments and on larger-scale models, which are representative of real-world production scenarios.

\subsection{Ablation Study}
To further dissect the sources of \algname 's performance advantages, we designed an ablation study to isolate the contribution of our core Stateful-MLFQ scheduling algorithm from the adaptive KV cache management policy. To achieve this, we deployed Stateful-MLFQ and all baseline scheduling algorithms on top of the native vLLM framework, which uses a PagedAttention memory manager without cache swapping or discarding.

\begin{figure}[h]
    \centering 
    \captionsetup[subfigure]{skip=1pt}
    \begin{subfigure}[b]{0.9\columnwidth}
        \centering
        \includegraphics[width=\columnwidth]{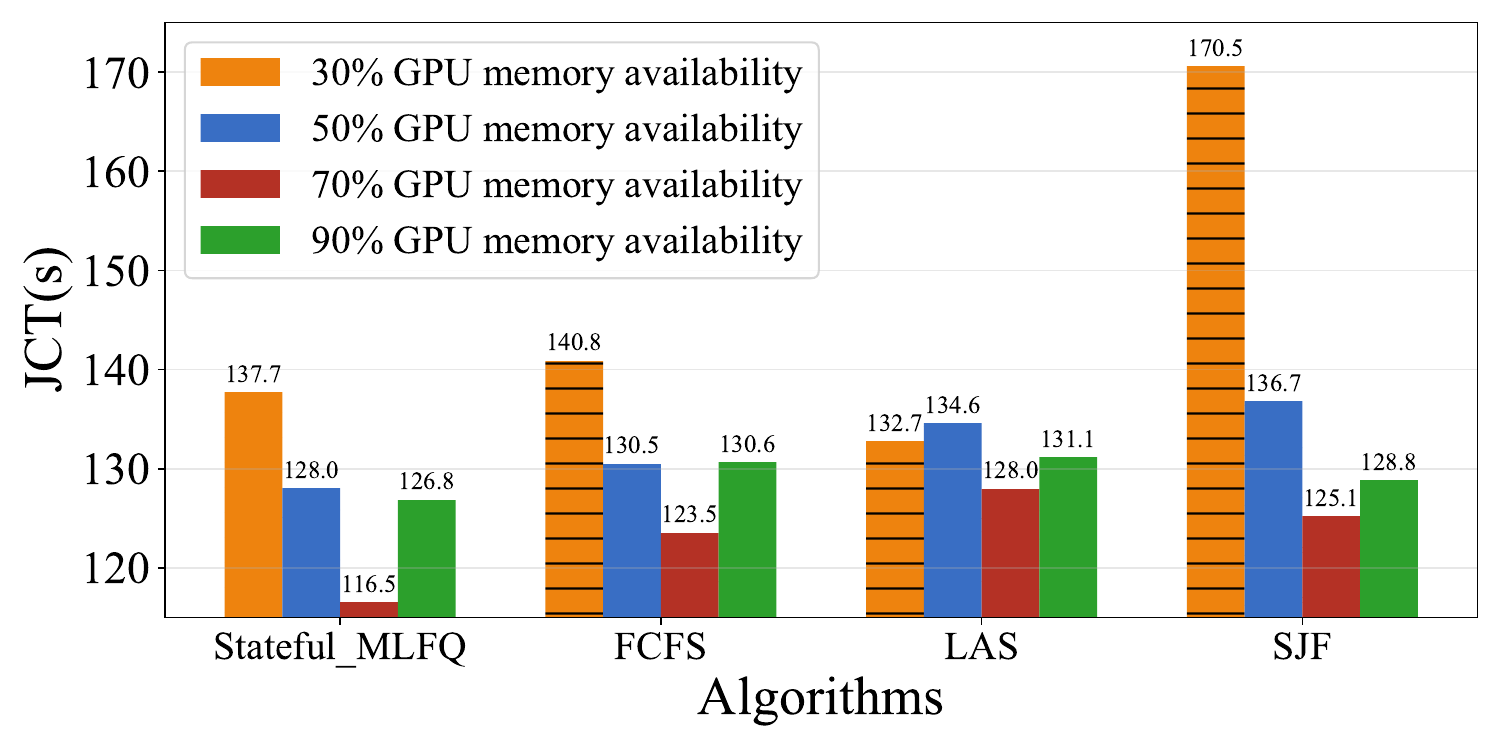} 
        \caption{GPT-J model}
        \label{fig:ablation_gptj}
    \end{subfigure}
    \hfill 
    \begin{subfigure}[b]{0.9\columnwidth}
        \centering
        \includegraphics[width=\columnwidth]{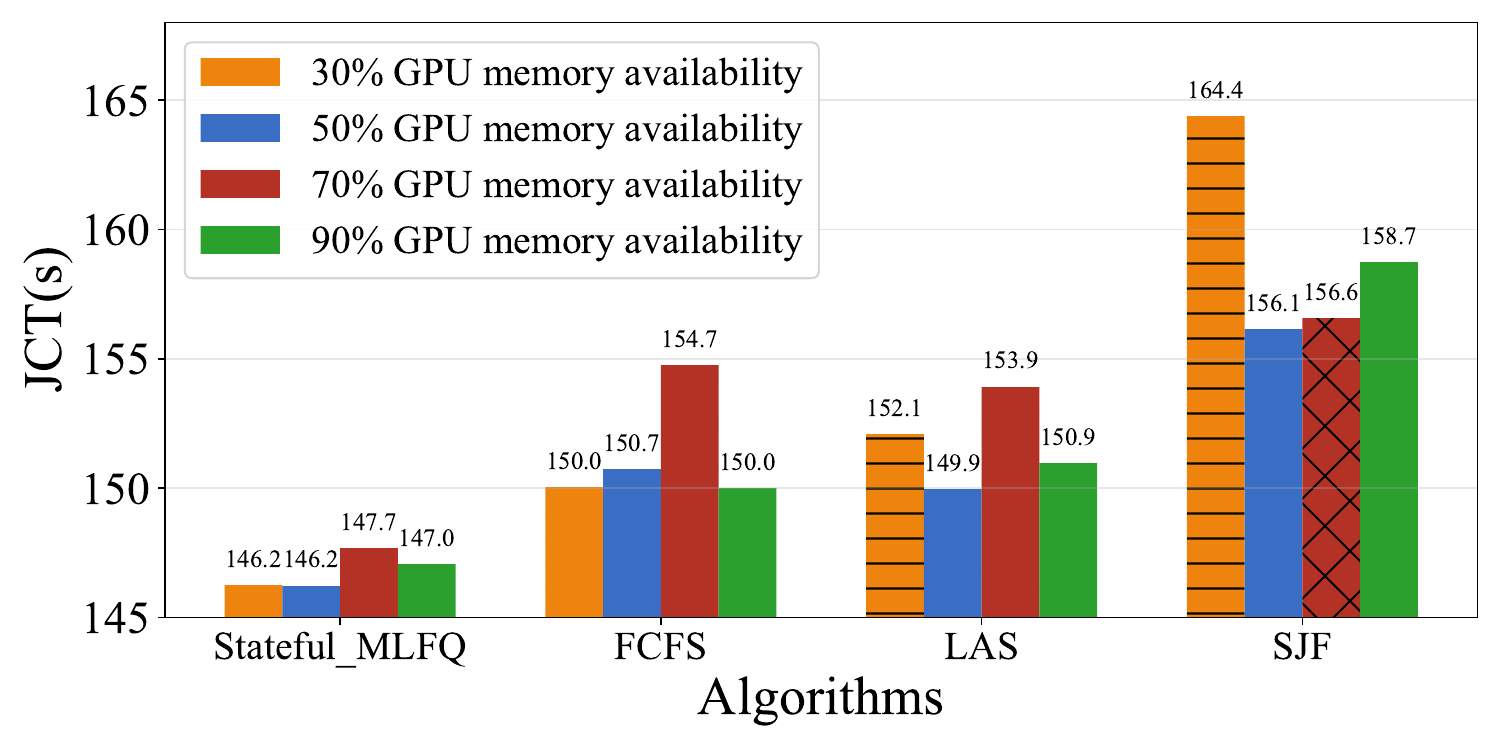} 
        \caption{Vicuna-13B model}
        \label{fig:ablation_vicuna}
    \end{subfigure}

    \caption{Ablation study comparing the performance of different scheduling algorithms on the (a) GPT-J and (b) Vicuna-13B models at a fixed load of QPS=3.}
    \label{fig:ablation_combined} 
    \Description{This is an example image showing ...}
\end{figure}
The results at a fixed system load of QPS=3 are shown in Figure~\ref{fig:ablation_combined}. On the GPT-J model (Figure~\ref{fig:ablation_gptj}), Stateful-MLFQ demonstrates the best performance across all memory configurations. Its advantage is particularly evident under high memory pressure. For instance, at 30\% memory availability, Stateful-MLFQ's JCT of 137.66s is 2.3\% lower than FCFS, and a significant 19.3\% lower than SJF. This shows that even without our adaptive cache manager, the lifecycle-aware scheduling logic can effectively mitigate queuing delay and head-of-line blocking.

We further validated the algorithm's effectiveness on the larger Vicuna-13B model (Figure~\ref{fig:ablation_vicuna}). In the most memory-constrained (30\%) scenario, Stateful-MLFQ's average JCT was 146.24s, yielding performance improvements of 2.5\%, 3.8\%, and 11.0\% compared to FCFS, LAS, and SJF, respectively. This result confirms the robust advantage of our state-aware scheduling policy under high memory pressure on large-scale models. In summary, the ablation study proves that Stateful-MLFQ is a potent scheduling algorithm in its own right, exhibiting good generality and stability across different models and resource configurations.

\subsection{Sensitivity, Overhead, and Stability}

We conducted further analyses to evaluate the robustness of our hyperparameter choices and the practical overhead of our system. A parameter sensitivity study confirmed that our default-chosen thresholds for both the aging mechanism and the MLFQ token costs are effective and near-optimal. Furthermore, we quantified the computational overhead of the \algname scheduler, finding it to be negligible, accounting for only 0.0006\% of the average JCT in a high-load scenario.

We also analyzed system stability by measuring the degradation of JCT as QPS increases. Under high memory pressure (30\% availability), the JCT of baseline methods like vLLM-SJF increased by as much as 51.9\% when moving from low to high load, whereas \algname's JCT increased by only 16.2\%. This demonstrates \algname's superior stability and robustness in contentious environments. \textbf{The detailed results and corresponding figures for these analyses are provided in Appendix~B and ~C.}

\section{Related Work}
\label{sec:related_work}
The optimization of LLM inference serving is a rapidly evolving field. Our work is positioned within the emerging domain of advanced scheduling for multi-stage agentic workflows, establishing a distinct research direction that differs from prior work in memory and batching optimizations and scheduling for one-shot queries.

\paragraph{\textbf{Memory and Batching Optimizations.}}
A lot of research aims to mitigate the large KV cache overhead to improve GPU throughput. Innovations in memory management, such as vLLM's PagedAttention~\cite{ref26}, LightLLM's token-level management~\cite{ref27}, and S³'s pre-allocation scheme~\cite{ref28}, focus on finer-grained control. Other approaches include OLLA~\cite{steiner2022olla}, which optimizes array lifetime and location to cut memory usage without model changes, and FlashAttention~\cite{dao2022flashattention}, an IO-aware exact attention algorithm that reduces memory reads/writes for faster training and longer sequences. Concurrently, advanced batching techniques like ORCA's continuous batching~\cite{ref31} and the split-and-merge in Sarathi~\cite{ref32} and DeepSpeed-FastGen~\cite{ref33} optimize request grouping. Batching innovations also cover DVABatch~\cite{cui2022dvabatch}, introducing diversity-aware multi-entry multi-exit batching for DNN serving, and FlexGen~\cite{sheng2023flexgen}, enabling high-throughput LLM inference on a single GPU via memory aggregation and compression. While foundational, these methods mainly optimize individual computational segments and not the inter segment scheduling in agentic workflows with long I/O phases.

\paragraph{\textbf{Scheduling Policies.}}
Another line of work focuses on scheduling policies. For single-request optimization, FastServe~\cite{ref34} uses job-level preemption to prioritize short queries. REEF~\cite{han2022microsecond} achieves microsecond-scale preemption for low-latency GPU-accelerated\\ DNN inference. In multi-tenant scenarios, VTC~\cite{ref35} employs a cost function to ensure fairness. Other directions include disaggregation, with systems like Splitwise~\cite{ref37}, TetriInfer~\cite{ref38}, and DistServe~\cite{ref39} separating prefill and decode stages for better load balancing. Model parallelism is leveraged by AlpaServe~\cite{li2023alpaserve} for statistical multiplexing, improving latency under bursty workloads. Llumnix~\cite{sun2024llumnix} introduces dynamic scheduling to handle heterogeneous requests and tail latencies. However, these schedulers target traditional, one-shot LLM queries. Infercept~\cite{ref29} and LAMPS~\cite{ref30} pioneer offloading KV cache during I/O waits to improve memory utilization. Autellix~\cite{ref36} first proposed a request-level scheduling policy (LAS) for agentic tasks.
In contrast to Autellix's retrospective policy, \algname is the first to introduce proactive scheduling that leverages predictions of both compute and I/O durations to optimize for the global JCT.
\section{Conclusion}
\label{sec:conclusion}
This paper focuses on the efficiency challenges posed by agentic workflows emerging from tool-augmented large language models (LLMs). Such workflows often involve frequent external API calls, particularly Web-based retrieval and service access. In these scenarios, the alternating pattern of computation and I/O commonly leads to significant end-to-end latency. To address this, we model the request lifecycle with the global objective of minimizing average job completion time, and prove that the scheduling problem is NP-hard. Building on this, we propose \algname, an inference serving engine tailored for agentic workflows. At its core lies a state-aware multi-level feedback queue (Stateful-MLFQ) scheduling algorithm that dynamically adjusts priorities and extends optimization from individual segments to the entire lifecycle. Experiments across diverse workloads and models demonstrate that \algname substantially reduces average JCT, validating its effectiveness and robustness in Web-augmented inference scenarios.

\clearpage
\bibliographystyle{plain}
\bibliography{references}
\appendix
\section{Problem Complexity}
\label{app:np_proof_detail}

We prove that the scheduling problem is NP-hard by reduction from the classic NP-hard problem of minimizing the total completion time on a single machine with precedence constraints, denoted as $1|prec|\sum C_j$.

\begin{proof}
An instance of $1|prec|\sum C_j$ consists of a set of jobs $J = \{J_1, J_2, ..., J_n\}$, where each job $J_k$ has a processing time $p_k$, and a precedence relation \textit{prec}.

Given an arbitrary instance of $1|prec|\sum C_j$, we construct an instance of our LLM agent scheduling problem as follows. We consider a simplified version of our problem where the batch size is one, meaning the GPU can only process one computational segment at a time.

\begin{enumerate}
    \item \textbf{Decomposition}: First, we decompose the precedence graph of jobs into a set of disjoint chains (e.g., $J_a \to J_b \to J_c$) and isolated jobs (e.g., $J_d$).

    \item \textbf{Mapping Construction}:
    \begin{itemize}
        \item For each chain of jobs, like $J_a \to J_b \to J_c$, we create a single parent request $R_i$ composed of a sequence of segments $\langle S_i^1, S_i^2, S_i^3 \rangle$.
        \item For each isolated job $J_d$, we create a single-segment request $R_j = \langle S_j^1 \rangle$.
    \end{itemize}

    \item \textbf{Service Time Assignment}: The processing time $p_k$ of each job $J_k$ is mapped to the \textbf{computational service time} of its corresponding segment $S_k$. For the purpose of this reduction, we assume the API call time for all constructed segments is zero. Specifically:
    \begin{itemize}
        \item For the request $R_i$ derived from the chain $J_a \to J_b \to J_c$, we set: $T_{comp}(S_i^1) = p_a$ and $T(A_i^1) = 0$; $T_{comp}(S_i^2) = p_b$ and $T(A_i^2) = 0$; and so on.
        \item For the request $R_j$ derived from the isolated job $J_d$, we set $T_{comp}(S_j^1) = p_d$ and $T(A_j^1) = 0$.
    \end{itemize}
\end{enumerate}

This construction is completed in polynomial time. By setting the API call times to zero, the dependency between segments $S_i^j$ and $S_i^{j+1}$ becomes an immediate precedence constraint that perfectly mimics the $J_k \to J_l$ relationship in the original problem. A valid schedule for the constructed requests is thus a permutation of the computational tasks $\{T_{comp}(S_1), ..., T_{comp}(S_n)\}$ that respects the intra-request ordering, which directly corresponds to a valid schedule for the jobs $\{J_1, ..., J_n\}$.

The completion time of a job $J_k$ ($C_k$) is equivalent to the completion time of the computational part of its corresponding segment $S_k$. Therefore, minimizing the sum of job completion times $\sum C_k$ is equivalent to minimizing the sum of segment completion times. As minimizing the total (or average) Job Completion Time (JCT) of all requests requires minimizing the sum of its underlying segment completion times, a solution to our problem provides a solution to the $1|prec|\sum C_j$ problem.

Since an arbitrary instance of the NP-hard problem $1|prec|\sum C_j$ can be reduced in polynomial time to a special case of our LLM agent scheduling problem (where batch size is 1 and API times are 0), our general problem is also NP-hard.
\end{proof}

\section{Parameter Sensitivity Analysis}
To validate the rationality and robustness of the key hyperparameter choices in our proposed method, this section conducts a sensitivity analysis on the aging threshold and the queue cost thresholds. All experiments were performed on the GPT-J model with 50\% memory availability.

\subsection{Analysis of Aging Threshold}
\begin{figure}[!h]
    \centering
    \includegraphics[width=0.9\linewidth]{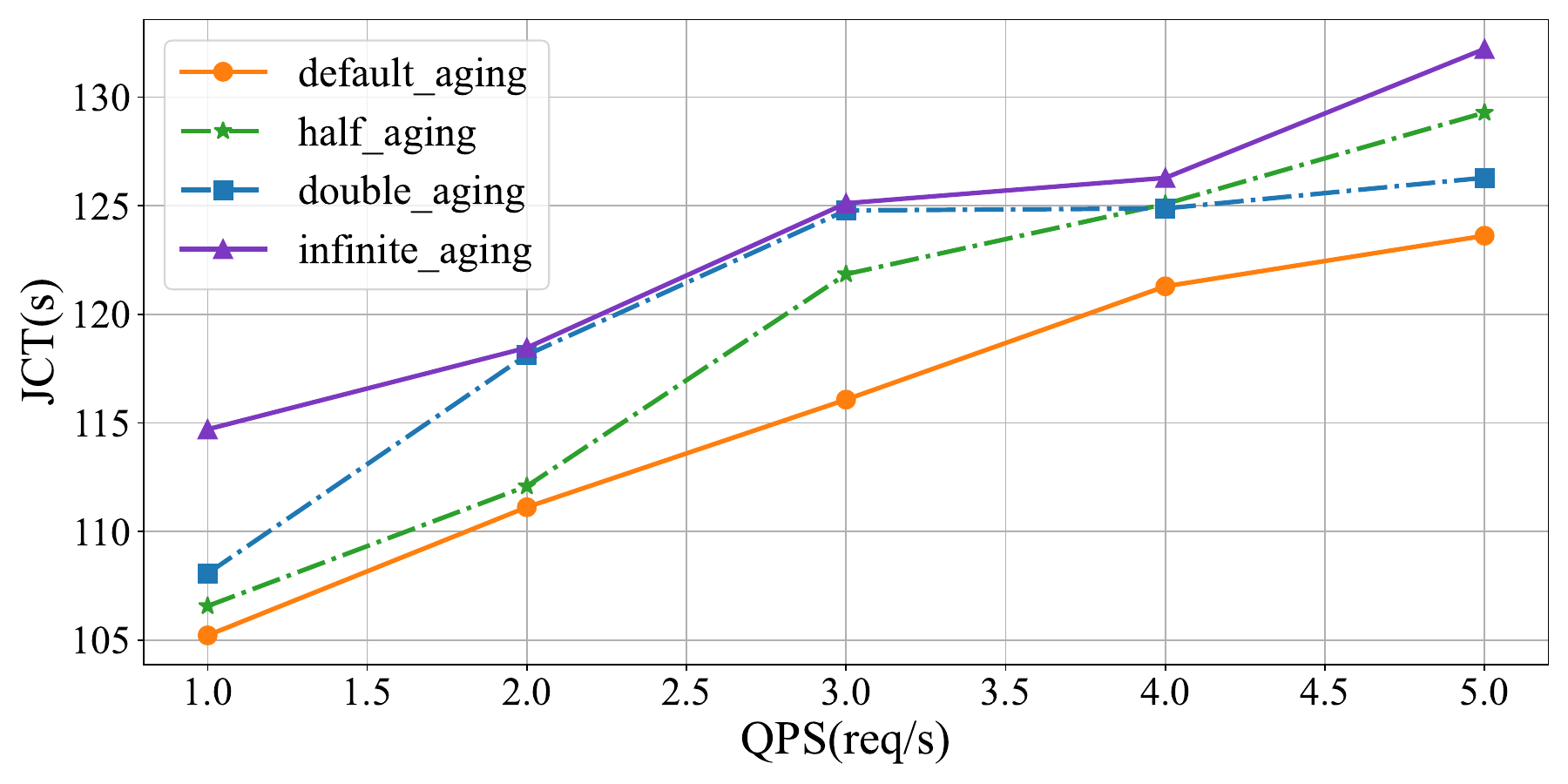} 
    \caption{Impact of the aging threshold on performance under different system loads.}
    \label{fig:sensitivity_aging}
    \Description{This is an example image showing ...}
\end{figure}

The aging mechanism is key to ensuring fairness in the Stateful-MLFQ algorithm. To investigate the impact of its core parameter, the aging threshold, we conducted a series of sensitivity experiments, setting the threshold (response ratio) to its default value of 5, a halved value of 2.5, a doubled value of 10, and infinity (disabling the mechanism). The results in Figure~\ref{fig:sensitivity_aging} show that disabling the aging mechanism performed the worst at all tested QPS loads, with its JCT being up to 7\% higher than the default configuration. This confirms the critical role of the anti-starvation mechanism. Furthermore, the results show that halving or doubling the threshold from the default value both led to an increase in average JCT, proving that an optimal parameter range exists.

\subsection{Analysis of Queue Cost Thresholds}
\begin{figure}[!ht]
    \centering
    \includegraphics[width=0.9\linewidth]{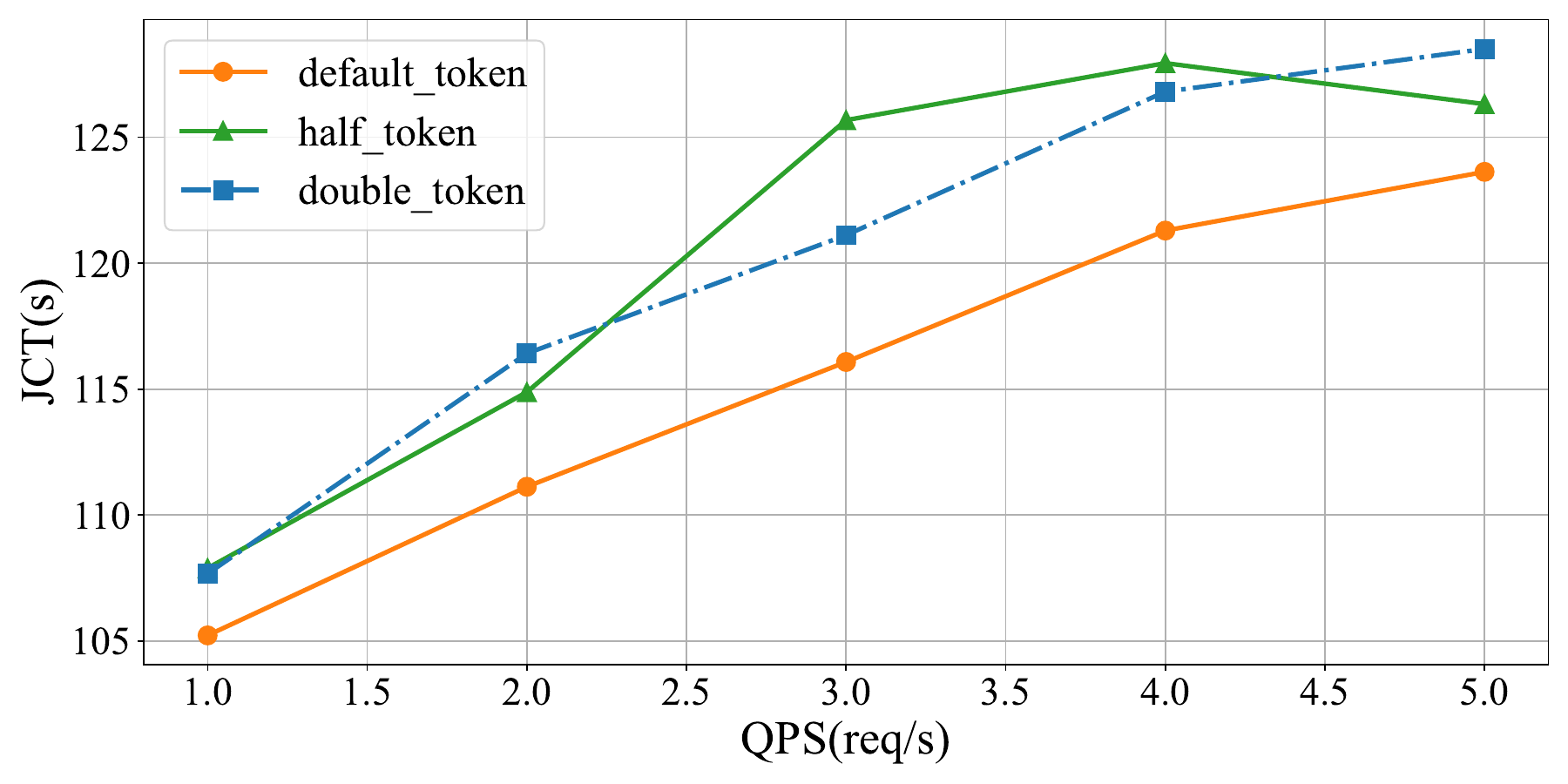} 
    \caption{Impact of queue cost thresholds on performance under different system loads.}
    \label{fig:sensitivity_token}
    \Description{This is an example image showing ...}
\end{figure}

In the Stateful-MLFQ algorithm, queue migration thresholds are determined by the computational cost (token count) of a segment. The default thresholds for the six queues were set to [128, 256, 384, 512, 640]. To examine the sensitivity of this parameter, we halved and doubled the thresholds while keeping other configurations constant. The results in Figure~\ref{fig:sensitivity_token} show that the default configuration maintained optimal performance across all loads. Setting the thresholds too low caused medium-length segments to be demoted prematurely, while setting them too high weakened the protection for I/O-intensive requests.

\section{System Overhead and Stability Analysis}
To quantify the computational overhead of the Stateful-MLFQ scheduling logic itself, we tested the duration of a single scheduling decision. Under a typical high-load scenario (GPT-J 6B model, 50\% memory availability, QPS=5), the average single scheduling decision time for \algname was 0.132 milliseconds. Given that a complete request in the workload averages 5.17 segments, the total scheduling overhead for a full request lifecycle is approximately 0.68 milliseconds. Compared to the average JCT of 123.6 seconds in this scenario, the total computational overhead of the scheduler itself is negligible, accounting for only 0.0006\%.

The stability of \algname is best demonstrated by its performance under high-load and memory-constrained conditions, as shown in the performance evaluation (Figure~\ref{fig:gptj_perf}). Stability can be measured by the steepness of the JCT curve as QPS increases.
In the most contentious scenarios (e.g., 30\% memory availability on GPT-J, Figure~\ref{fig:gptj_30}), baseline methods exhibit high instability. For instance, the JCT of vLLM-SJF skyrockets from 116.05s at QPS=1 to 176.34s at QPS=5, a 51.9\% increase, indicating severe performance degradation under load. In contrast, \algname's JCT curve remains significantly flatter, rising from 105.77s to only 122.88s, an increase of just 16.2\%.
This demonstrates that \algname possesses superior load adaptability and scheduling stability. By effectively mitigating resource contention and head-of-line blocking through its state-aware policies, it maintains a predictable and robust performance profile, which is a critical requirement for deploying LLM agent services in production environments with fluctuating loads and stringent resource constraints.
\end{document}